\documentclass{amsart}
\usepackage{amsopn}

\usepackage{bbold}
\usepackage{amsmath}
\usepackage{amscd}
\usepackage{amsthm}
\usepackage{mathrsfs }
\usepackage{amsfonts}

\usepackage{caption}
\usepackage{amssymb}
\usepackage[square,sort,comma,numbers]{natbib}

\usepackage{graphicx, color}

\usepackage{subcaption}
\usepackage{array}

\begin{document}

\title{Network Distance Based on Laplacian Flows on Graphs}
\author{Dianbin Bao}
\address{316 Sutherland Building, 1600 Woodland Road, Abington, PA,19001}
\email{dub835@psu.edu}

\author{Kisung You}
\address{B02 Hayes-Healy Center,
	Notre Dame, IN 46556-5641}
\email{kyou@nd.edu}

\author{Lizhen Lin}
\address{Hurley 152A, Notre Dame, IN 46556 }
\email{lizhen.lin@nd.edu}
\begin{abstract}
Distance plays a fundamental role in measuring similarity between objects. Various visualization techniques and learning tasks in statistics and machine learning such as shape matching, classification, dimension reduction and clustering  often rely on some distance or similarity measure. It is of tremendous importance to have a distance that can incorporate the  underlying structure of the object.  In this paper,  we focus on proposing such a distance between network objects.  Our key  insight is to define a distance based on the long term diffusion behavior of the whole network. We first introduce a dynamic system on graphs called Laplacian flow. Based on this Laplacian flow, a new version of diffusion distance between networks is proposed. We will demonstrate the utility of the distance and its advantage over various existing distances through  explicit examples.  The distance  is also applied to subsequent  learning tasks such as clustering network objects.
\end{abstract}

\maketitle
\section{Introduction}

A network is a representation of relations between objects and arises naturally in characterizing phenotypes of complex data. Due to its flexibility in representing the underlying structure of data, networks have presented their significance in a variety of scientific fields from biology and neurosicence (brain and biological  networks) to social science (social networks), to name just a few. This has necessitated immense developments in theory, methodologies and algorithms over the last few decades for inference of a network. For instance, there are many models for clustering  nodes within a network such as stochastic blockmodels \cite{karrer_stochastic_2011, holland_stochastic_1983}, spectral clustering \cite{von_luxburg_tutorial_2007}, modularity optimization \cite{newman_modularity_2006} and so on.

Recently, there has been an emerging strong need of a framework to make inference on a population of network objects. For example, the Human Connectome Project \cite{van_essen_human_2012} hosts brain imaging data of more than one thousand subjects where an individual's neural system is represented as a network object. Datasets of such type pose many difficulties in providing convincing answers to questions like clustering, centrality, hypothesis test, and others at the population level.  There has been some recent work on inferences of a population of networks based the notion of Fr\'echet means \citep{frechet, linclt} with applications to hypothesis testing with network data  \cite{ginestet2017, unlabeled-18}.  \cite{NIPS2017_lin} proposes to cluster network objects based a mixture of graphon model.  Answering many other questions of this type starts from defining a descriptor to measure similarity between  network objects. Surprisingly little attention has been shown but we will introduce some of the previous works in the next section. One notable work in the Computer Science literature is graph kernel \cite{vishwanathan_graph_2010} which computes kernel similarity matrices.

The rest of the paper is organized as follows. In Section \ref{review}, we give a brief description of distance measures between network objects. Based on Laplacian flow, we propose a novel distance measure, characterize its properties, and provide an efficient numerical scheme in Section \ref{work}. Our simulation in Section \ref{simulation} supports our new proposal to outperform incumbent metrics.

% For instance, in a sensor network, one would like to know whether or not there is a hole in the coverage of all sensors. To obtain global information of a network it often requires a thorough understanding of how objects are combined locally in a network. The machinery of algebraic topology, which provides a tool to compute global invariant in terms of local data, is now incorporated into statistics and coined the name topological data analysis. See \cite{Edelsbrunner2002} and \cite{carlsson2005} for more on this topic.
%	

%Sometimes a data point contains more information than a vector. For instance, in image recognition, the image at the level of pixels can be restored as a matrix. If we use only the vector space structure, then each coordinate of a vector is on an equal footing, therefore inter-relation between pixels is lost. Another example concerns networks. A network $G$ can be determined by its adjacent matrix $A$. Again we lose global information if we view $A$ as a vector.  This is the reason that we study flow dynamics on an object to obtain its global structure information. 

%In computer version, a dataset may be a shape of a particular object and we may use graph as an approximation and study the distance between graphs to distinguish different objects. 

\section{Related Work}	\label{review}

Let $G$ be a network or graph with $n$ nodes with its adjacency matrix $A$, which is an $n\times n$ matrix. For a binary network, $A_{ij} = 1$ if node $i$ and $j$ has an observed edge between two nodes and $A_{ij}=0$ otherwise for $1\leq i,j \leq n$. The graph Laplacian of a graph $G$ is defined as $\mathcal{L} = D-A$, where $D$ is a degree matrix such that $D_{ii} = \sum_{j}A_{ij}$ for $1\leq i \leq n$ and $D_{ij}=0$ otherwise \cite{von_luxburg_tutorial_2007}. 

Several measures have been proposed to describe dissimilarity based on direct observables of the network. One simple way is to count the number of matching edges from two networks \cite{hamming_error_1950}, the popular Hamming distance. In \cite{wilson_study_2008}, Wilson and Zhu suggested to use the Euclidean distance between the spectra of two adjacency matrices. From a network-theoretic perspective, Roy et al. \cite{roy_modeling_2014} claimed the discrepancy of node-defined centrality measures be a candidate for dissimilarity measure. 

The graph Laplacian has been known as an approximation of the Laplace-Beltrami operator on smooth manifold underlying observed objects \cite{gine_empirical_2006}. Since $\mathcal{L}$ contains geometric and topological information of the data via its spectrum, many strategies have been proposed. 

Since the graph Laplacian matrix is symmetric and positive-semidefinite, eigenvalues of $\mathcal{L}$ are nonnegative real numbers. Jakobson and Rivin exploited such phenomenon by defining the distance measure by taking normalized sum of squared differences for top eigenvalues \cite{jakobson_extremal_2002}. Instead of using eigenvalues directly, some chose to compute the disparity of two approximated distributions of spectrum. Ipsen and Mikhailov, in \cite{ipsen_evolutionary_2002}, suggested to apply kernel density estimation by convolving narrow Lorentz distributions with computed eigenvalues, while  Fay et al. employed discrete histogram through binning \cite{fay_weighted_2010}.

Recently, an interesting work adopted diffusion dynamics on a graph to characterize and distinguish networks. The work is based on the manifold learning method called Diffusion Maps \cite{coifman_geometric_2005}, where the distance between two nodes takes all possible paths in between into account across different timescale. This implies that each network and its topological properties can be well presented by the diffusion process. In \cite{hammond2013graph}, Hammond et al. adopted such idea to define graph diffusion distance that measures dissimilarity of two networks and showed it is indeed a metric.

\section{Proposed Work}	\label{work}

Suppose we have two graphs $G_1$ and $G_2$ with the same number of $N$ nodes. Let $\textbf{c}^1(t)=(c^1_1(t),\dots, c^1_N(t))$ be a time-dependent vector of functions associated with the nodes of $G_1$. Similarly we define $\textbf{c}^2(t)$ for $G_2$. The Laplacian flow is a dynamic system defined in a coordinate-wise manner by 
\begin{equation}\label{dynamic}
\dot{c_i}(t)=\sum_{j\sim i}(c_j(t)-c_i(t)),
\end{equation}
where the sum runs over all nodes adjacent to $i$, and a compact expression for equation \eqref{dynamic} is
\begin{equation}\label{dynamicvector}
\dot{\textbf{c}}(t)=-\mathcal{L}\textbf{c}(t).
\end{equation}
Given an initial condition $\textbf{c}(0)$, we can solve the  system to obtain an analytic solution,
\begin{equation}\label{solution}
\textbf{c}(t)=\exp(-t\mathcal{L})\textbf{c}(0).
\end{equation}
Since the eigenvalues of $\mathcal{L}$ are nonnegative, the solution will converge to the projection of $\textbf{c}(0)$ to the kernel of $\mathcal{L}$. Now we give the same initial condition for the two graphs $G_1$ and $G_2$ so that 
\begin{equation}
\textbf{c}^i(t)=\exp(-t\mathcal{L}_i)\textbf{c}(0)
\end{equation}
for $i=1,2$. Graph diffusion distance in \cite{hammond2013graph} is defined as maximal discrepancy on a family of distance measures across different time points,
\begin{equation}\label{hammond}
d_{GDD}(G_1,G_2)=\max_{t}||\exp(-t\mathcal{L}_1)-\exp(-t\mathcal{L}_2)||_F,
\end{equation}
where the subscript $F$ means Frobenius norm for a matrix.

\subsection{ Definition of the Network Flow Distance (NLD)}

We study the difference between diffusion processes  at the nodes $[i^1]$ in $G_1$ and $[i^2]$ in $G_2$ using $\dot{c}^1_i(t)-\dot{c}^2_i(t)$ for {\em various initial conditions}. Define 
\begin{equation}\label{distance}
d_i:=\sum_{\textbf{c}(0)}\int_0^\infty |\dot{c}_i^1(t)-\dot{c}_i^2(t)|dt,
\end{equation}
where in the sum $\textbf{c}(0)$ runs through standard basis vectors $e_j=(0,\cdots,1,\cdots,0)$ for all $j\neq i$. Although the definition uses an improper integral, one can see the convergence without difficulty. Moreover, the integrand at $t=0$ is given by the absolute value of the $i$-th component of $\mathcal{L}_1\textbf{c}(0)-\mathcal{L}_2\textbf{c}(0)$. When $\textbf{c}(0)$ runs through basis vectors $e_j$'s for all $j\neq i$, we find that the integrand coincides with the Hamming distance of the $i$-th row of the adjacent matrices. 
Then we define the {\em network flow distance (NLD)} between two graphs as 
\begin{equation}\label{distancesum}
d_{NLD}(G_1,G_2):=\sum_{i=1}^nd_i.
\end{equation}
From definitions \eqref{distance} and \eqref{distancesum} it is straightforward to check that $d_{NLD}$ satisfies the well known axioms of a distance metric, i.e., 
\begin{enumerate}
	\item[(i)] $d_{NLD}$ is symmetric,
	\item[(ii)]  $d_{NLD}(G_1,G_2)\geq 0$ and $d_{NLD}(G_1,G_2)=0$ if and only if $G_1$ and $G_2$ are identical,
	\item[(iii)] $d_{NLD}(G_1,G_3)\leq d_{NLD}(G_1,G_2)+d_{NLD}(G_2,G_3)$.
\end{enumerate}
The definition $d_{GDD}$ has a similar flavor as $d_{NLD}$ in nature by incorporating the diffusion behavior of a whole network but there are some key differences. One drawback of the definition of $d_{GDD}$ is that the maximum may occur at  a different time for a different pair of graphs, which results in mismatching behavior in the context of a large group of graphs. We integrate the distance between a pair of nodes between two graphs with respect to time. In practice, we can truncate at proper $T_{\max}$. Due to the exponential decay of the integrand in \eqref{distance}, $T_{\max}$ can be chosen properly according to one's desired precision.  Another advantage is that we removed the diagonal terms so that we characterize a node in a network entirely through its environment and not  nodes itself. 
%In addition, $\tilde d$ does not take into account any initial conditions which might impact the diffusion behavior of the network. 
Moreover, $d_{GDD}$  fails to capture the long term behavior of the diffusion process by considering the discrepancy of the diffusions of the networks at a single time point $t$.  Our simulation study confirms that the network flow distance outperforms $d_{GDD}$ significantly in distinguishing networks under various settings and in using the distances for clustering network objects.

\subsection{An efficient computation scheme}

In this subsection, we discuss our method for computing the  distance defined in the last subsection. In particular, we propose a computation scheme that enables fast computation of our network flow distance $d_{NLD}$. 
For convenience, let $f_i(t):=c^1_i(t)-c_i^2(t)$. From equation \eqref{solution}, we know that $|\dot{f}_i(t)|$ decays exponentially to $0$ as $t\rightarrow\infty$. Truncating the improper integral in equation \eqref{distance} at a properly chosen $T_{\max}$ yields the approximation
\begin{equation}
\int_{0}^{\infty}|\dot{f}_i(t)|dt\approx\int_{0}^{T_{\max}}|\dot{f}_i(t)|dt.
\end{equation}
Using finite difference method, we have the following approximation
\begin{equation}\label{appro}
\int_{0}^{T_{\max}}|\dot{f}_i(t)|dt\approx\sum_{k=1}^{N}|f_i(t_k)-f_i(t_{k-1})|,
\end{equation}
where $t_0=0$ and $t_N=T_{\max}$.
With simple arithmetrics, we know the right hand side of \eqref{appro} has cancellations due to the alternating nature of the terms $f_i(t_k)-f_i(t_{k-1})$ when $f_i(t)$ is monotone on an interval $I$. Note that $f_i(0)=0$ and for connected graphs $f_i(t)\rightarrow 0$ for $t\rightarrow \infty$, then we see that $\int_{0}^{\infty}|\dot{f}_i(t)|dt$ is determined by all extreme values of $f_i(t)$. It is interesting to compare with definition \eqref{hammond}. In equation \eqref{hammond} the $\max$ is taken globally for all nodes with respect to time, while in definition \eqref{distancesum}, we take sum of extreme values for each individual nodes. 

From the original definition of the Laplacian flow in equation \eqref{dynamicvector}, we see that using $f_i(t)$ instead of $\dot{f}_i(t)$ in \eqref{appro} reduces the multiplication by graph Laplacians, which is crucial since iterative multiplications by graph Laplacians can be computationally expensive. Define 
\begin{equation}
A(t,\mathcal{L}_1,\mathcal{L}_2)=\exp(-t\mathcal{L}_1)-\exp(-t\mathcal{L}_2).
\end{equation}
It is well known that a graph Laplacian $\mathcal{L}$ is symmetric and positive
semidefinite so that we have the following spectral decomposition:
\begin{equation}
\mathcal{L}=\Lambda D\Lambda^{T}. 
\end{equation}
Then we have 
\begin{equation}
\exp(-t\mathcal{L})=\Lambda \exp(-tD)\Lambda^{T}.
\end{equation}

For a matrix $M$, we define $g$ as the sum of absolute values of the off-diagonal entries of $M$, i.e.,
\begin{equation}
g(M)=\sum_{i\neq j}|M_{ij}|.
\end{equation}
By equations \eqref{distance},\eqref{distancesum} and \eqref{appro}, we have
\begin{equation}
d_{NLD}(G_1,G_2)\approx\sum_{i=1}^{N}g(A(t_i,\mathcal{L}_1,\mathcal{L}_2)-A(t_{i-1},\mathcal{L}_1,\mathcal{L}_2)).
\end{equation}
In the next section we will provide simulation examples in which our definition gives stronger and more precise cluster structure than that obtained using $d_{GDD}$ or the Hamming distance or the Frobenius norm distance between corresponding Laplacians of networks defined by $$d_F(G_i,G_j):=||\mathcal{L}_i-\mathcal{L}_j||_F.$$

	\section{Simulation Study }\label{simulation}
	
	%5and Data Analysis, find a real data set??
	
	In this section, we demonstrate the success of our distance using several examples in a simulation study. It can be shown that our distance can detect distances between certain networks while the popular Hamming distance or Frobenius distance between graph Laplacians can not.  We then apply the distance to clustering network objects based on a spectral clustering algorithm. 
	
	\subsection{Distance between networks with one edge deletion}
	
\begin{figure}[!tbp]
		\centering
		%\begin{minipage}[b]{0.5\textwidth}
			\includegraphics[width=0.3\textwidth]{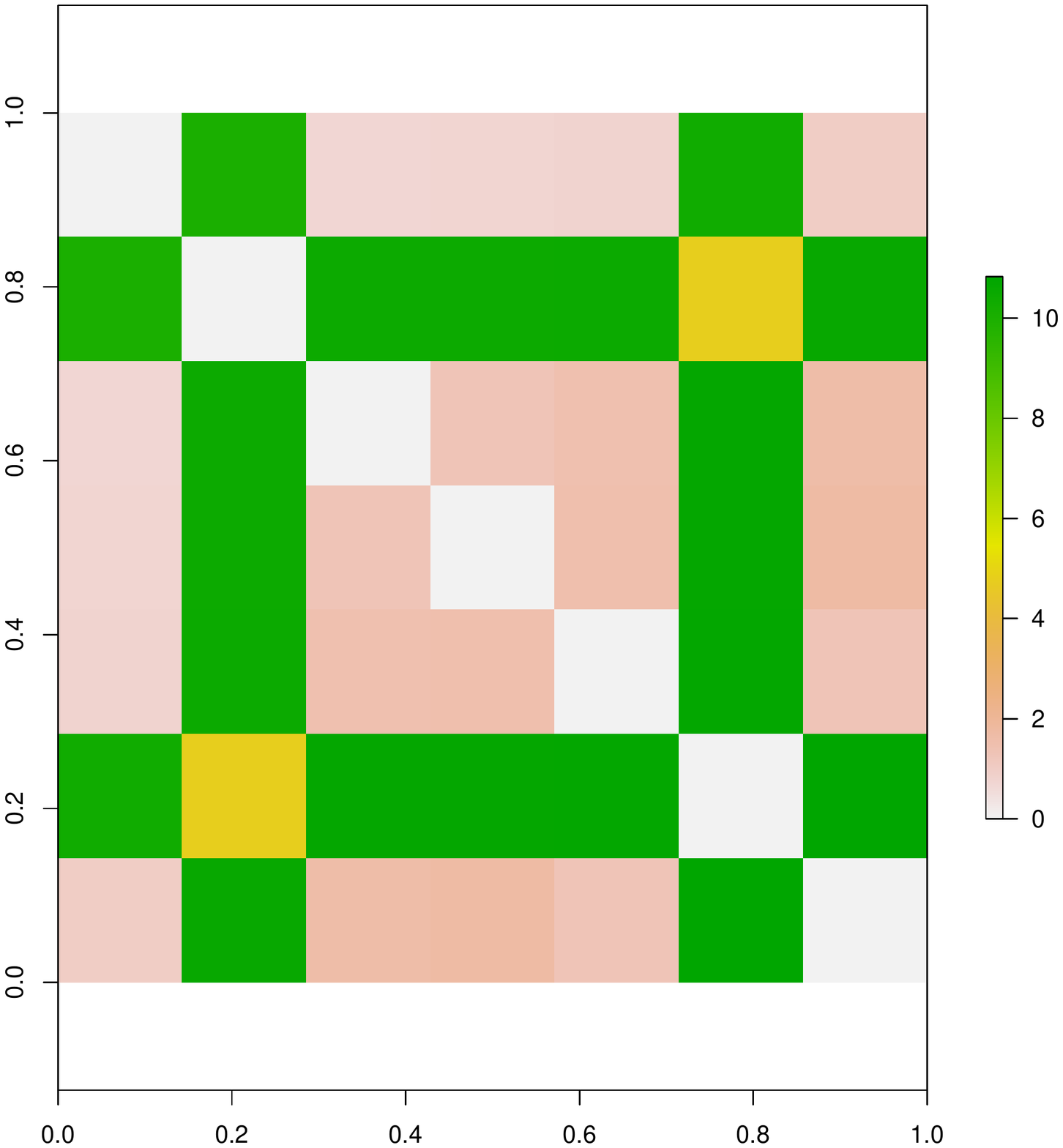}
		\caption{Network flow distance $d_{NLD}$.}
			\label{matrix2}
		\end{figure}

\begin{figure}
		\centering
			\includegraphics[width=0.3\textwidth]{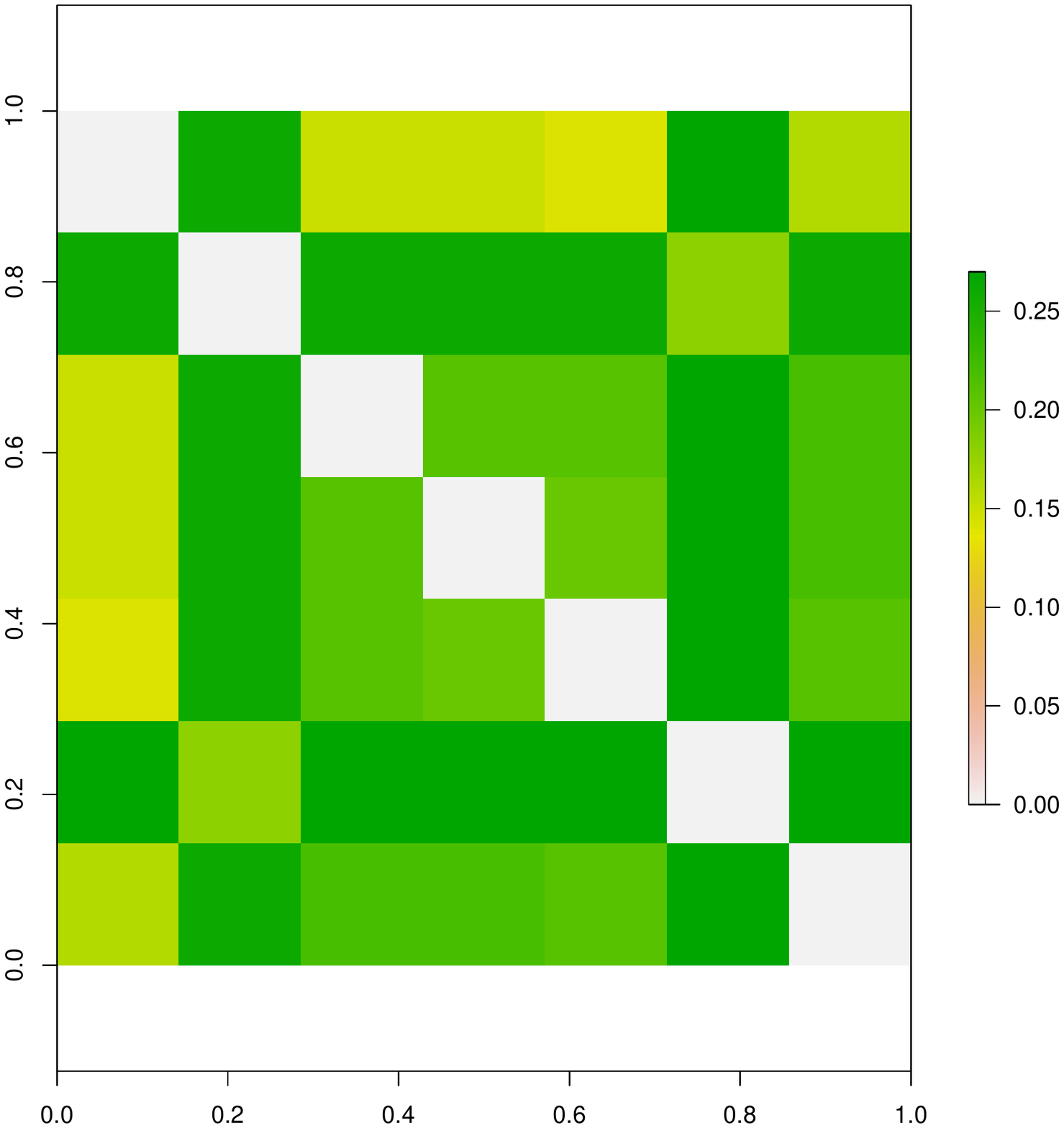}
		\caption{Network diffusion distance $d_{GDD}$.}
			\label{matrix1}
		\end{figure}

	Let $G_1$ be a graph with 20 nodes distributed equally to form two communities $C_1$ and $C_2$. We generate $G_1$ from a stochastic block model (SBM) with  edges between two nodes in $C_1$ (resp. $C_2$) with probability $P_{11}=0.75$ (resp. $P_{22}=0.6$) and generate inter-community edges with probability $P_{12}=0.04$. We use a uniform distribution to generate entries of the adjacent matrices with the probability above. The graph $G_1$  generated using $\textsf{R}$ with the above parameters has two bridges between its two communities $C_1$ and $C_2$. $G_2$ and $G_6$ are obtained from $G_1$ by removing one of the bridges. The other graphs are obtained from $G_1$ by removing a within-community edge. Since a bridge in general plays a more important role in a network, we expect that $G_2$ and $G_6$  have larger distances to $G_1$ than other graphs. Our numerical computation plotted in Figures \ref{matrix2} and  \ref{matrix1} shows that this is indeed the case. We take $T_{\max}=40$ and use 1200 sample points in our computation. The computation process takes only seconds on a MAC  desktop with 3.6 GHz Intel Core i7 Processor. Note that the Hamming distance between  these graphs is 1 or 2 and it completely fails to tell the difference between a bridge edge and a within community edge. 
	
	%5\textcolor{red}{Dianbin, please add a sentence or two on the $d_F$ distance between their corresponding Laplacians, we have space for that.}
	
	\begin{figure}[!tbp]
		\centering
		\begin{minipage}[b]{0.3\textwidth}
			\includegraphics[width=\textwidth]{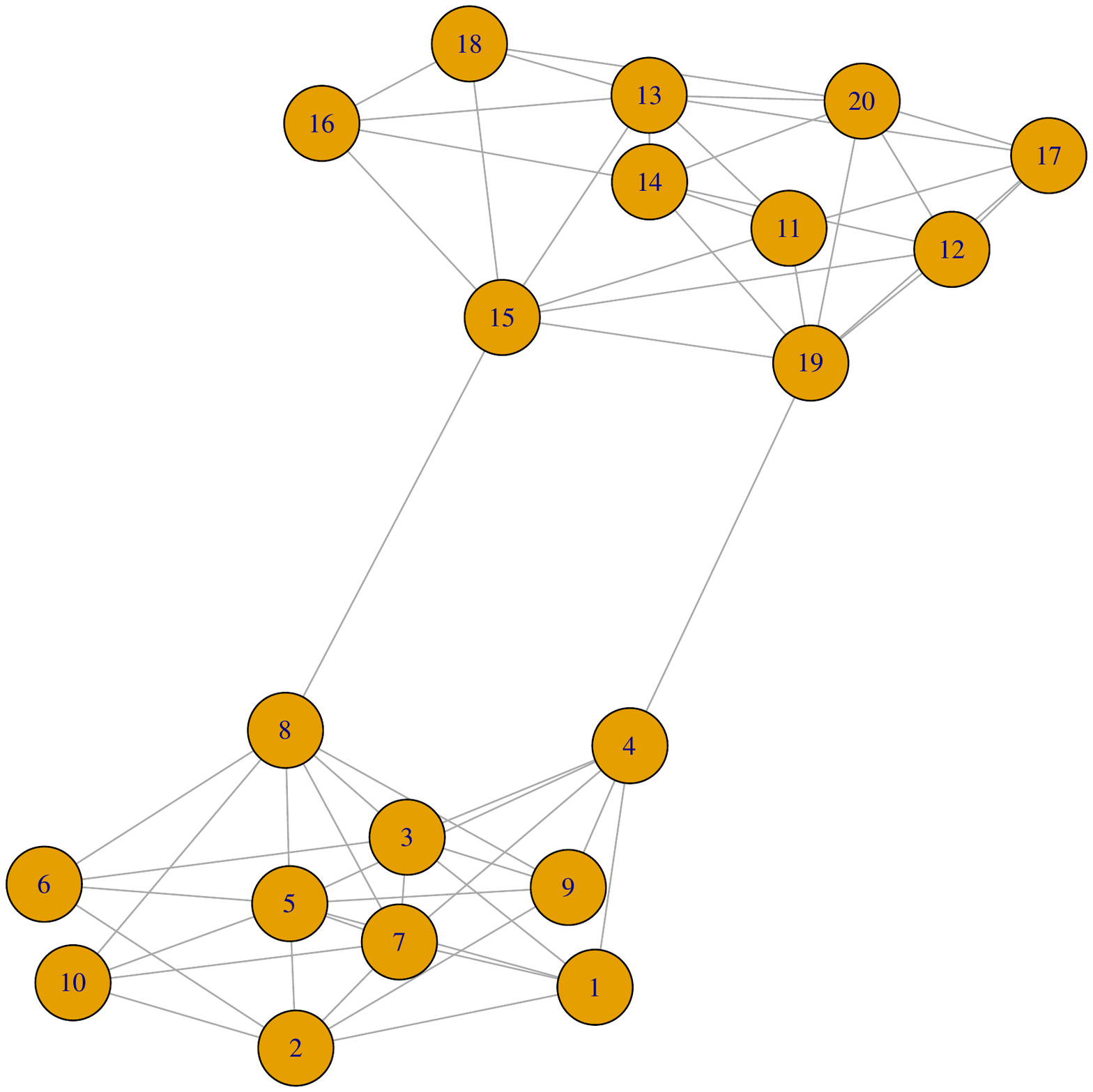}
			\caption{G1.}
			\label{G1}
			
		\end{minipage}
		%\hfill
		\begin{minipage}[b]{0.3\textwidth}
		\includegraphics[width=\textwidth]{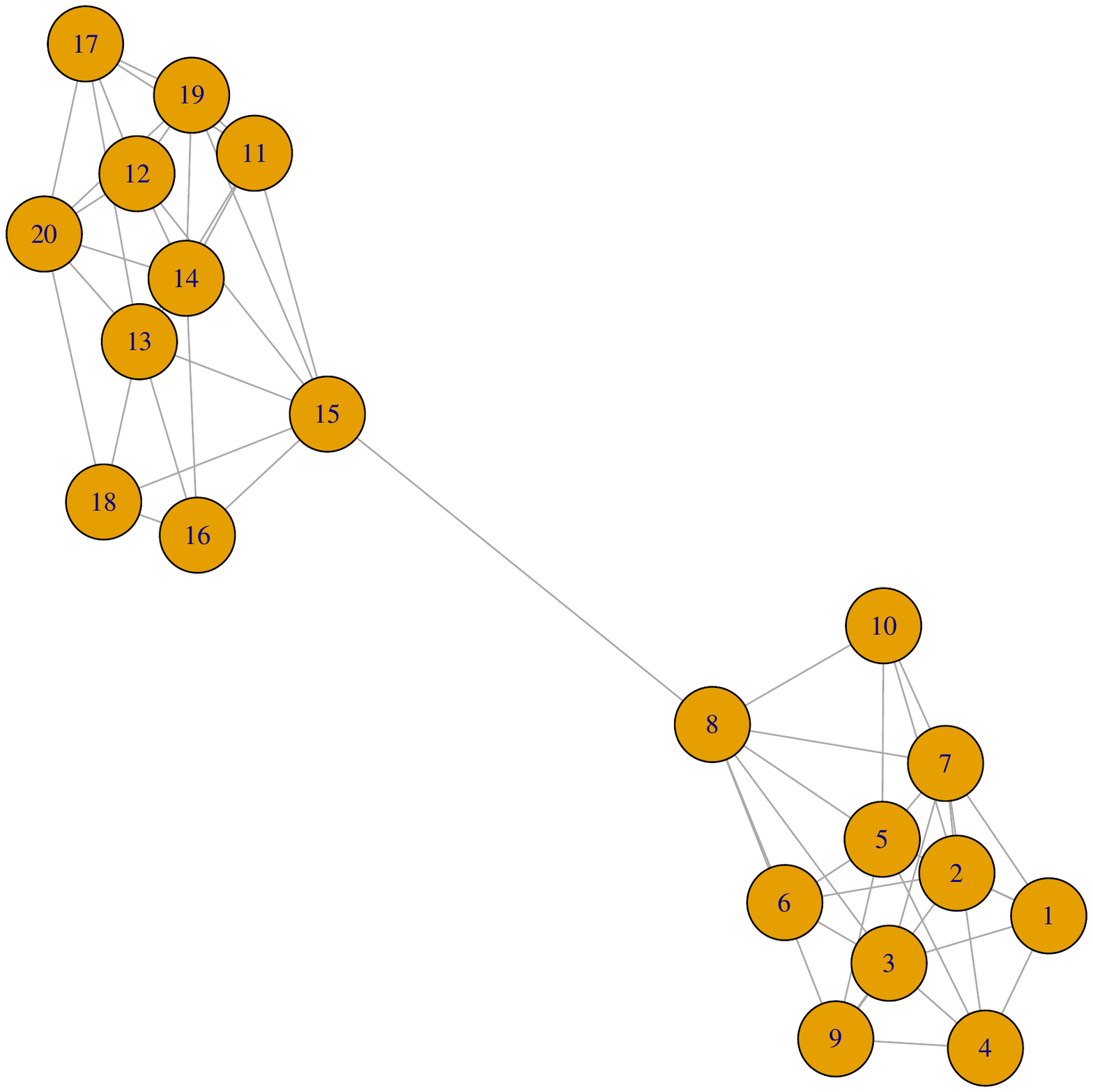}
			\caption{G2.}
			\label{G2}
			
		\end{minipage}
		%\hfill
		\begin{minipage}[b]{0.3\textwidth}
		\includegraphics[width=\textwidth]{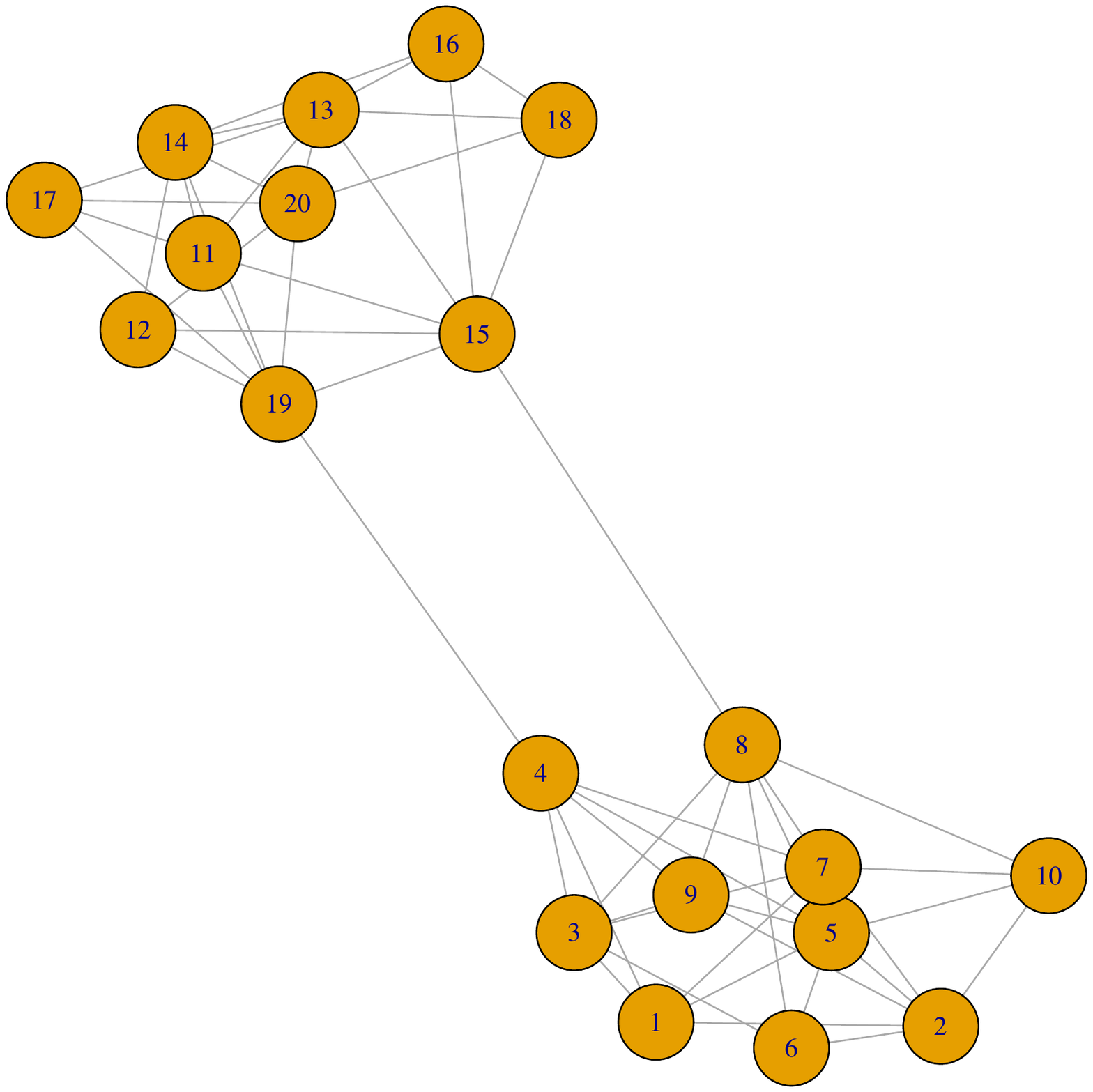}
			\caption{G3.}
			\label{G3}
		\end{minipage}
	\end{figure}

	%\hfill
		%\begin{minipage}[b]{0.5\textwidth}
	%\hfill
		
		\begin{figure}
		%\begin{minipage}[b]{0.5\textwidth}
		\centering
			\includegraphics[width=0.3\textwidth]{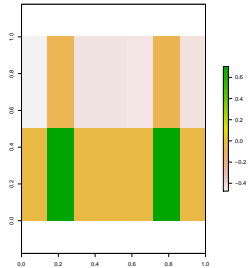}
		\caption{Two eigenvectors of $d_{NLD}$.}
			\label{eigenvector}
		%\end{minipage}
	\end{figure}

	Now we compare the distance matrices obtained using $d_{GDD}$, the maximal distance defined in Section 3 with our distance $d_{NLD}$. The distance matrix of $d_{GDD}$ is plotted in Figure \ref{matrix1}. 
%	For instance if we compute the relative difference of distances between $G_1$,  $G_2$ and $G_3$ using $d_{GDD}$, we will get
%	$$\frac{d_{GDD}(G_1,G_2)-d_{GDD}(G_1,G_3)}{d_{GDD}(G_1,G_3)}=\frac{0.26-0.15}{0.15}\approx0.73,$$
%	while using $d_{NLD}$ we have
%	$$\frac{d_{NLD}(G_1,G_2)-d_{NLD}(G_1,G_3)}{d_{NLD}(G_1,G_3)}=\frac{10.02-0.68}{0.68}\approx13.74.$$
	We see that in this case our proposed distance outperforms the distance $d_{GDD}$ significantly.
	The cluster structure in Figure \ref{matrix2} is so strong that the desired cluster structure can be obtained directly using the $k$-means algorithm. Indeed, we apply the $k$-means algorithm to any column or row of the distance matrix in Figure \ref{matrix2} with cluster number 2 and find that the output cluster vector by \textsf{R} separates $G_2$ and $G_6$ from the other graphs. For this example, we can also obtain our desired cluster structure using the spectral clustering algorithm. Indeed, if we define a similarity matrix $S$ by 
	\begin{equation}\label{similarity}
	S_{ij}=\exp(-d_{NLD}(G_i,G_j)/\sigma),
	\end{equation}
    where $\sigma$ is the standard deviation of $S_{ij}$, then we obtain the two eigenvectors of $S$ with largest two eigenvalues in Figure \ref{eigenvector}.
   	
   	%\lstinputlisting[float=h,frame=tb,caption=Clustering vectors,label=eigenvector ]{eigenvectors.txt}
%   	
%   	\begin{figure}[!tbp]
%		\centering
%		\begin{minipage}[b]{0.3\textwidth}
%			\includegraphics[width=\textwidth]{EigenvectorNLD.pdf}
%		\caption{Two eigenvectors of $d_{NLD}$.}
%			\label{eigenvector}
%		\end{minipage}
%	\end{figure}
%\begin{table}
%\caption{Clustering vectors}
% \label{eigenvector}
%\begin{lstlisting}
%[1] -0.474 -0.088 -0.439 -0.436 -0.440 -0.084 -0.430
%[2] -0.057  0.701 -0.054 -0.053 -0.055  0.703 -0.054
%\end{lstlisting}
% \end{table}
%   	

    Applying the $k$-means algorithm to the two vectors in Figure \ref{eigenvector} with cluster number $k=2$, then we obtain our desired cluster structure. However, it will be another case if we use Hamming distance instead. Indeed, the Hamming distance is given by $d_{\text{Hamming}}(i,j)=2$ for $i,j\neq 1$ and $i\neq j$ and $d_{\text{Hamming}}(1,i)=1$ for $i\neq 1$. 
	If we compute the corresponding similarity matrix using Hamming distance and apply the spectral clustering algorithm, then we find that the output cluster vector given by \textsf{R} is $(2,1,2,2,2,2,2)$.  The cluster vector singles out $G_2$ from the other graphs.  However, from our construction we know that $G_2$ and $G_6$ are more similar. If we use distance  $d_{F}$, then we compute the distance matrix
	using that $||\mathcal{L}_i-\mathcal{L}_j||_{F}=2\sqrt{2} $ for $i\neq j$ and $i,j\neq 1$ and that $||\mathcal{L}_1-\mathcal{L}_j||_{F}=2$ for $j\neq 1$. If we apply spectral clustering algorithm to the similarity matrix $S_F$ for the distance $d_F$, then we obtain cluster vector  $(1,1,1,1,2,2,1)$ using \textsf{R}. In this case, graphs $G_5$ and $G_6$ are put in one cluster, which is different from our desired cluster containing $G_2$ and $G_6$. 
	The failure of Hamming distance $d_{\text{Hamming}}$ and Frobenius distance  $d_F$  is not  surprising since both distances assign the same weight to two apparently different types of edges.
	
\begin{figure}[!tbp]
		\centering
		%\begin{minipage}[b]{0.5\textwidth}
			\includegraphics[width=0.4\textwidth]{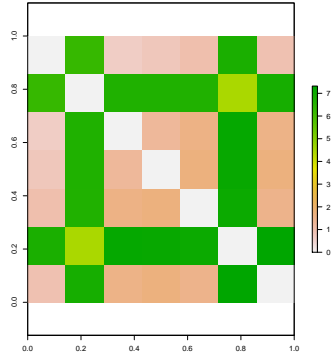}
	\caption{Network flow distance matrix with one more bridge}
	\label{distance1}
	\end{figure}

\begin{figure}
	\centering
		%\end{minipage}
		%\hfill
		%\begin{minipage}[b]{0.5\textwidth}
			\includegraphics[width=0.4\textwidth]{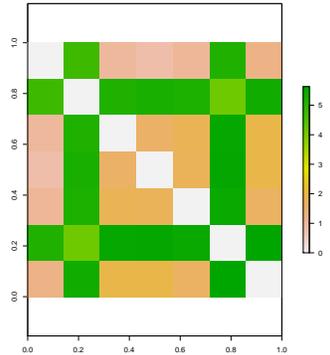}
	\caption{Network flow distance matrix with two more bridges}
	\label{distance2}
	\end{figure}

		Another feature that the network flow distance outperforms the Hamming distance and the distance $d_F$ is that the network flow distance is \emph{floating according to the dynamics of a network graph. }
	In a dynamical network, the role of an edge may change in various ways.  For instance, in the game of Go, the value of a previous move is floating as the game tree develops.  Keep the notation of $G_i$ in the last example,	
but we now add more bridges to the graphs, then the  distances $d_{NLD}(G_1,G_2)$ and $d_{NLD}(G_1,G_6)$ are expected to decrease since the importance of one bridge is diluted. Indeed, see the distance matrices in Figure \ref{distance1} (resp. Figure \ref{distance2} ), which is obtained by adding one bridge $(1,11)$ (resp. two bridges $(1,11)$ and $(2,12)$) in $G_i$s.

%\lstinputlisting[float=h,frame=tb,caption=Diffusion distance matrix with one more bridge,label=distance1 ]{diffusiondistance1.txt}	
	\begin{figure}
		%\end{minipage}
	%\end{figure}
%\begin{figure}[!tbp]
		\centering
		%\begin{minipage}[b]{0.5\textwidth}
			\includegraphics[width=0.4\textwidth]{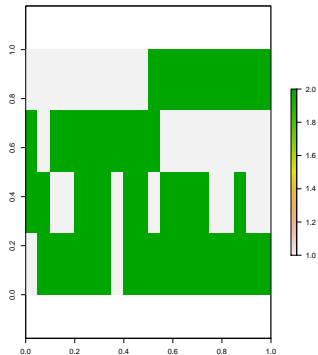}
		\caption{Cluster vectors (first two rows use $d_{NLD}$ and last two rows use $d_{GDD}$)}
			\label{clustervector}
		%\end{minipage}
	\end{figure}

%\begin{table}
%	\caption{Network flow distance matrix with one more bridge}
%	\label{distance1}
%\begin{lstlisting}
%     [,1] [,2] [,3] [,4] [,5] [,6] [,7]
%[1,] 0.00 6.24 0.65 0.80 1.01 6.76 0.95
%[2,] 6.24 0.00 6.68 6.70 6.65 4.46 6.87
%[3,] 0.65 6.68 0.00 1.31 1.62 7.20 1.58
%[4,] 0.80 6.70 1.31 0.00 1.73 7.19 1.72
%[5,] 1.01 6.65 1.62 1.73 0.00 7.10 1.53
%[6,] 6.76 4.46 7.20 7.19 7.10 0.00 7.32
%[7,] 0.95 6.87 1.58 1.72 1.53 7.32 0.00
%\end{lstlisting}
%	\end{table}

	% \lstinputlisting[float=h,frame=tb,caption=Diffusion distance matrix with two more bridges,label=distance2 ]{Diffusiondistance2.txt} 

%\begin{table}
%\caption{Network flow distance matrix with two more bridges}
%\label{distance2}
%\begin{lstlisting}
%    [,1] [,2] [,3] [,4] [,5] [,6] [,7]
%[1,] 0.00 4.73 0.97 0.81 1.01 5.11 1.21
%[2,] 4.73 0.00 5.15 5.23 5.18 4.10 5.37
%[3,] 0.97 5.15 0.00 1.60 1.80 5.55 1.94
%[4,] 0.81 5.23 1.60 0.00 1.74 5.58 1.94
%[5,] 1.01 5.18 1.80 1.74 0.00 5.48 1.64
%[6,] 5.11 4.10 5.55 5.58 5.48 0.00 5.63
%[7,] 1.21 5.37 1.94 1.94 1.64 5.63 0.00	 	
%\end{lstlisting}
%\end{table}
	We see that $d_{NLD}(-,G_2)$ and $d_{NLD}(-,G_6)$ are decreasing as expected by comparing Figures \ref{matrix2}, \ref{distance1} and \ref{distance2}.
	%Indeed, using Hamming distance, we have $d_{\text{Hamming}}(1,j)=1$ for $j\neq 1$ and $d_{\text{Hamming}}(i,j)=2$ for $i\neq j$ and $i,j\neq 1$. 
	%Then the distance matrix is given by 
	%$$d_{\text{Hamming}}=\left(\begin{matrix}
	% 0 & 1 & 1 & 1 & 1 & 1 & 1 \\ 
	%   1 & 0 & 2 & 2 & 2 & 2 & 2 \\ 
	%   1 & 2 & 0 & 2 & 2 & 2 & 2 \\ 
	%  1 & 2 & 2 & 0 & 2 & 2 & 2 \\ 
	% 1 & 2 & 2 & 2 & 0 & 2 & 2 \\ 
	%   1 & 2 & 2 & 2 & 2 & 0 & 2 \\ 
	%   1 & 2 & 2 & 2 & 2 & 2 & 0 \\ 
	%  \end{matrix}\right)$$

	% The distance matrix for $d_F$ is given by 
	%$$2\left(\begin{matrix}
	% 0 & 1 & 1 & 1 & 1 & 1 & 1 \\ 
	%   1 & 0 & \sqrt{2} & \sqrt{2} & \sqrt{2} & \sqrt{2} & \sqrt{2} \\ 
	%   1 & \sqrt{2} & 0 & \sqrt{2} & \sqrt{2} & \sqrt{2} & \sqrt{2} \\ 
	%  1 & \sqrt{2} & \sqrt{2} & 0 & \sqrt{2} & \sqrt{2} & \sqrt{2} \\ 
	% 1 & \sqrt{2} & \sqrt{2} & \sqrt{2} & 0 & \sqrt{2} & \sqrt{2} \\ 
	%   1 & \sqrt{2} & \sqrt{2} & \sqrt{2} & \sqrt{2} & 0 & \sqrt{2} \\ 
	%   1 & \sqrt{2} & \sqrt{2} & \sqrt{2} & \sqrt{2} & \sqrt{2} & 0 \\ 
	%\end{matrix}
	%\right)$$
	%We use the similarity matrix $S_{F}(i,j):=\exp(-d_F(i,j)^2/2\sigma^2)$ with $\sigma$ be the standard deviation of the nonzero entries of $d_{F}$ and This again is not intrinsic due to the  permutation symmetry of $\{G_2,\cdots,G_7\}$.  
	% \newpage
	\subsection{Illustration of the distance matrices between a collection of graphs}\label{example2}
	In this example we use a different setting. We only fix bridges and generate within-community edges randomly with a fixed probability $p$. 
	For simplicity, we use a two-block model and each block has 10 nodes. The probability of an edge between two nodes within the same block is $p$ and the cross-block edges are fixed. We generate 20 graphs, in which the first 10 have 5 fixed bridges and the other 10 graphs have 10 fixed bridges. Our simulation shows that network flow distance $d_{NLD}$ outperforms the Frobenius norm distance $d_F$ and the diffusion distance $d_{GDD}$ in this scenario. In our simulation we choose $p=0.8$ , $T_{\max}=4$ and we use 400 sample points to estimate the integral in \eqref{distance}. Then we apply the $k$-means algorithm to the rows or columns of the distance matrix after replacing the diagonal terms by an average. Then we find that our network flow distance gives very precise cluster structure, which tells apart the two different ways for constructing the graphs. However, the cluster structure obtained using the Frobenius norm distance $d_F$ or diffusion distance $d_{GDD}$ is unreliable.

\begin{figure}[!tbp]
		\centering
		%\begin{minipage}[b]{0.5\textwidth}
			\includegraphics[width=0.4\textwidth]{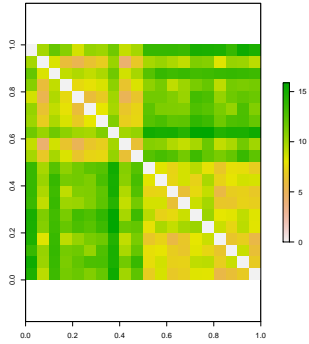}
		\caption{Network flow distance $d_{NLD}$.}
			\label{DMDF}
			\end{figure}

\begin{figure}
			\centering
			\includegraphics[width=0.4\textwidth]{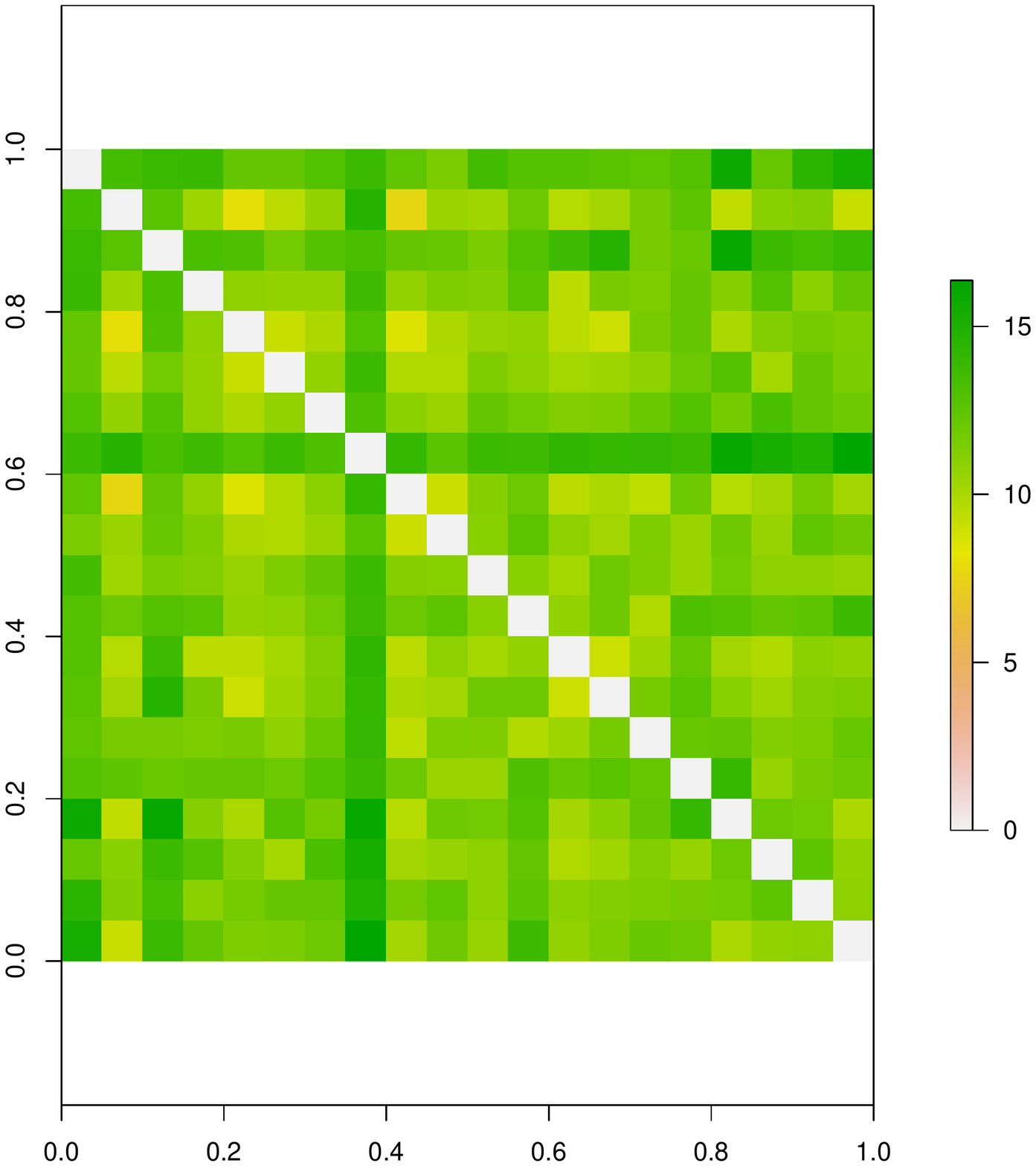}
		\caption{Frobenius norm distance $d_F$.}
				\label{DMFrobL}
				\end{figure}

\begin{figure}
		\centering
			\includegraphics[width=0.4\textwidth]{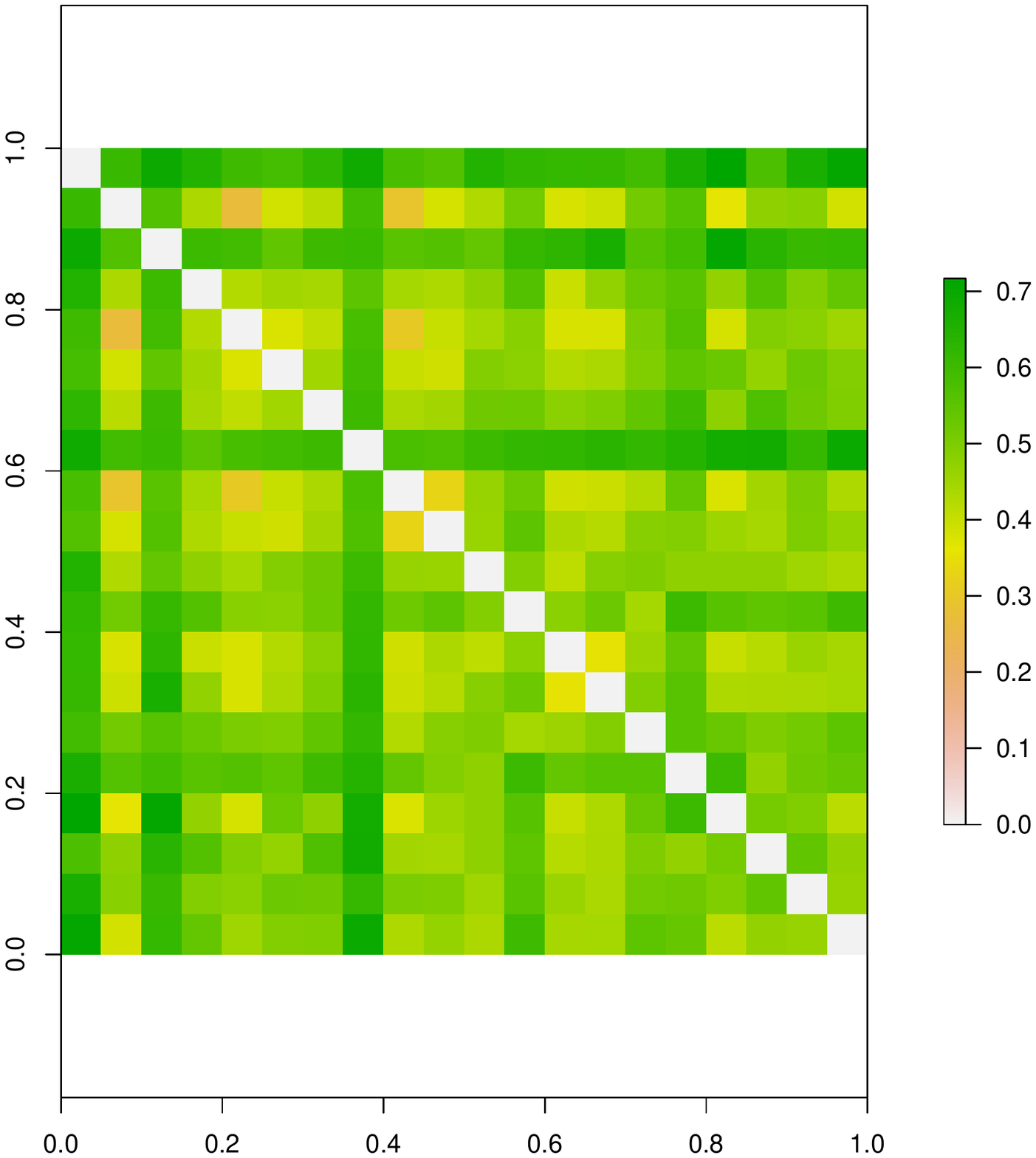}
			\caption{Diffusion distance $d_{GDD}$.}
			\label{DMDFold}
	%	\end{minipage}
	\end{figure}

	The distance matrices are plotted in Figures \ref{DMDF}, \ref{DMFrobL} and \ref{DMDFold}.  One sees that the network flow  distance $d_{NLD}$ shows the strongest cluster structure. We also noticed that  only the distance ${d_{NLD}}$ gives very precise cluster structure if we apply $k$-means algorithm to the distance matrices. We plot 4 cluster vectors in Figure \ref{clustervector}, where the first two vectors are obtained by applying $k$-means algorithm to the 1st and 11th rows of the distance matrix using $d_{NLD}$. The first gives a perfect cluster structure and the second has 2 misses ($G_2$,$G_{11}$). The 3rd and 4th cluster vectors are similarly obtained using $d_{GDD}$ and both vectors misclassified 8 out of 20 objects so that the cluster structure is poorly obtained in some sense.

%
%	\begin{table}
%\caption{Cluster vectors (first two rows use $d_{NLD}$ and last two rows use $d_{GDD}$)}
%\label{clustervector}		
%\begin{lstlisting}
%[1] 1 1 1 1 1 1 1 1 1 1 2 2 2 2 2 2 2 2 2 2
%[2] 2 1 2 2 2 2 2 2 2 2 2 1 1 1 1 1 1 1 1 1
%[3] 2 2 1 1 2 2 2 1 2 2 1 2 2 2 2 1 1 2 1 1
%[4] 1 2 2 2 2 2 2 1 2 2 2 2 2 2 2 2 2 2 2 2
%\end{lstlisting}
%		\end{table}
	
		%\end{minipage}
		%\hfill
		%\begin{minipage}[b]{0.5\textwidth}
		%\end{minipage}
		%\hfill
		%\begin{minipage}[b]{0.5\textwidth}
\begin{figure}[!tbp]
		\centering
		%\begin{minipage}[b]{0.5\textwidth}
			\includegraphics[width=0.4\textwidth]{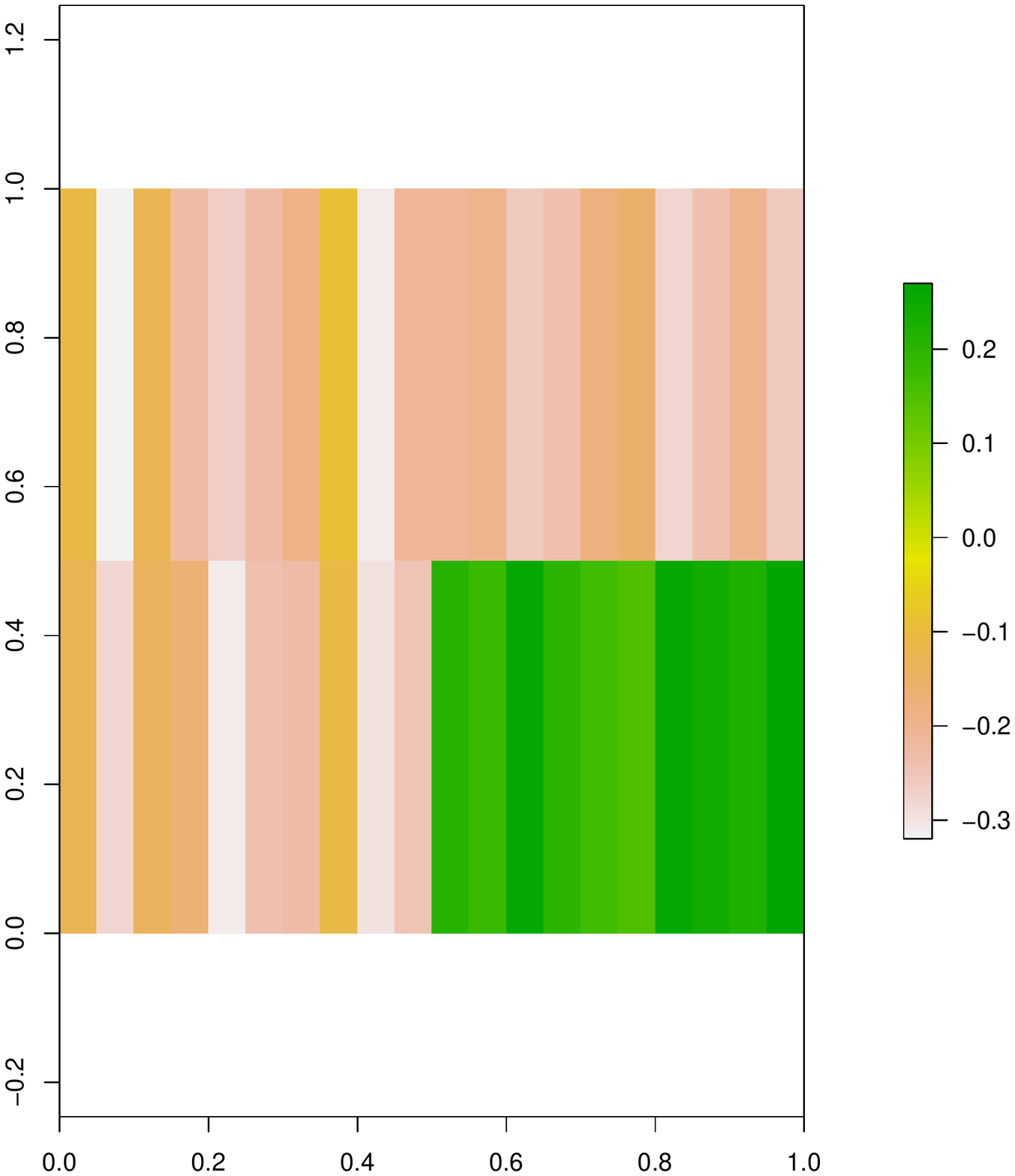}
		\caption{Eigenvectors corresponding to the largest two eigenvalues}
			\label{eigen2}
		%\end{minipage}
\end{figure}

In this case, we can also apply the spectral clustering algorithm. Again, we define similarity matrix $S$ by equation \eqref{similarity} and obtain the two eigenvectors of $S$ with the largest two eigenvalues using \textsf{R}. See Figure \ref{eigen2}.

%\begin{table}
%\caption{Eigenvectors corresponding to the largest two eigenvalues}
%\label{eigen2}
%\begin{lstlisting}
%[1] -0.11 -0.32 -0.13 -0.23 -0.27 -0.23 -0.19 -0.09 -0.31 -0.21
%-0.21 -0.20 -0.26 -0.24 -0.18 -0.16 -0.28 -0.24 -0.20 -0.26
%[2] -0.13 -0.28 -0.14 -0.17 -0.31 -0.24 -0.23 -0.11 -0.30 -0.25 
%0.21  0.18  0.26  0.20  0.17  0.15  0.26  0.24  0.22  0.27
%\end{lstlisting}
%\end{table}
%	
In particular the signs in the second eigenvector show a clear cluster structure. Apply $k$-means algorithm	to the two vectors in Figure \ref{eigen2} using \textsf{R}, then we obtain a perfect cluster structure.

%	
%	Another feature that the diffusion distance outperforms the Hamming distance is that the diffusion distance is floating according to the dynamics of a network graph. 
%	In a dynamical network, the role of an edge may change in various ways.  For instance, in the game of Go, the value of a previous move is floating as the game tree develops.  Keep the notation in the last example that $G_1$ is a graph obtained from a graph $G$ by deleting an edge $e$. Let $\mathcal{S}$ be a set of edges that has trivial intersection with the edge set $E$ of $G$ and a priori $e\notin \mathcal{S}$. Then it is possible that $d(G,G_1)\neq d(G\cup \mathcal{S},G_1\cup \mathcal{S})$. 
%	If we view $\mathcal{S}$ as a testing set, then from the pattern how 
%	$d(G\cup \mathcal{S},G_1\cup \mathcal{S})$ changes with respect to the testing set $\mathcal{S}$ we may expect to obtain some information of the edge $e$ in the network $G$. 
%	Continuing with the above example in Figure \ref{G1}-\ref{G3}, we now add more bridges to the graph, then the diffusion distances $d(G_1,G_2)$ and $d(G_1,G_6)$ are expected to decrease since the importance of one bridge is diluted.

\subsection{Clustering network objects from two stochastic block models}
\begin{figure}[!tbp]
	\centering
	%\begin{minipage}[b]{0.5\textwidth}
		\includegraphics[width=0.4\textwidth]{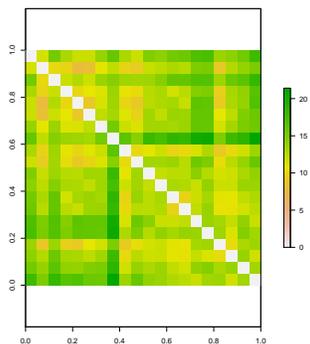}
		\caption{Network flow distance $d_{NLD}$.}
		\label{DMDFrandom}
		\end{figure}

\begin{figure}
	\centering
		\includegraphics[width=0.4\textwidth]{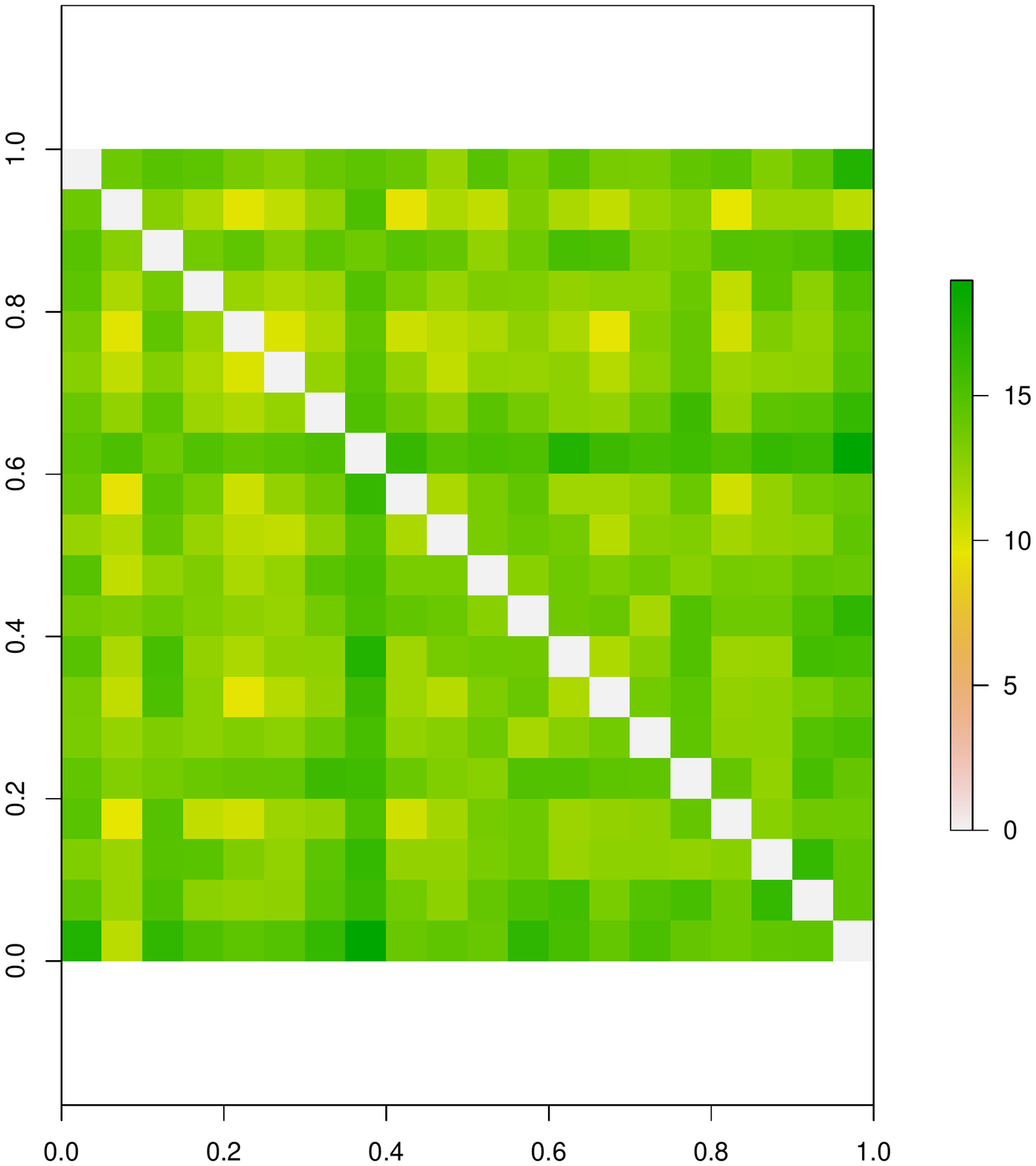}
		\caption{Frobenius norm distance $d_F$.}
		\label{DMFrobLrandom}
		\end{figure}

\begin{figure}
	\centering
		\includegraphics[width=0.4\textwidth]{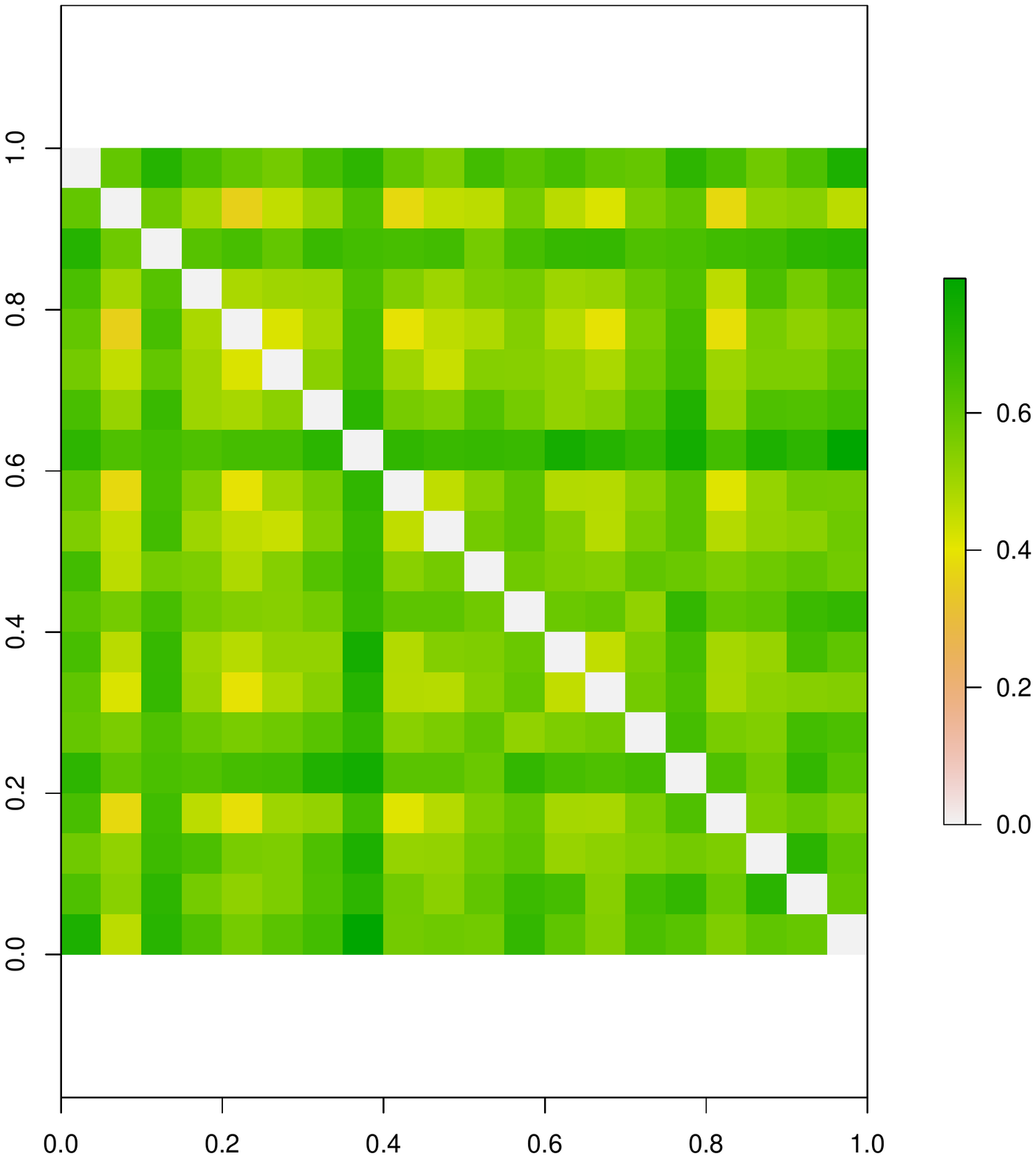}
		\caption{Diffusion distance $d_{GDD}$.}
		\label{DMDFoldrandom}
		\end{figure}

Now we test the difference between utilities of  various definitions of distance in an example of clustering network objects generated from two SBMs.  
%The  model similar to Example \ref{example2} in some sense.
 We generated 10 network objects from one stochastic block model (SBM) with within community link probabilities $P_{11}=P_{22}=0.8$  and  between community link probability $P_{12}=0.05$.  We generate another 10 networks from another SBM with the same within community link probabilities but the between community link probability is given by $2P_{12}$. The number of nodes for each graph is 20. 
%5he key difference is that the bridges are no longer fixed and instead we randomly generate bridges between communities for each of the first 10 graphs with probability $P_{12}=0.05$ and similarly for the other 10 graphs we generate bridges with probability $2P_{12}$. 
See the 2D plots for the distance matrices in Figures \ref{DMDFrandom}, \ref{DMFrobLrandom} and \ref{DMDFoldrandom}. 

	%\end{minipage}
	%\hfill
	%\begin{minipage}[b]{0.5\textwidth}
	%\end{minipage}
	%\hfill
	%\begin{minipage}[b]{0.5\textwidth}
	%\end{minipage}
	
%\end{figure}

\begin{figure}[!tbp]
		\centering
		%\begin{minipage}[b]{0.5\textwidth}
			\includegraphics[width=0.4\textwidth]{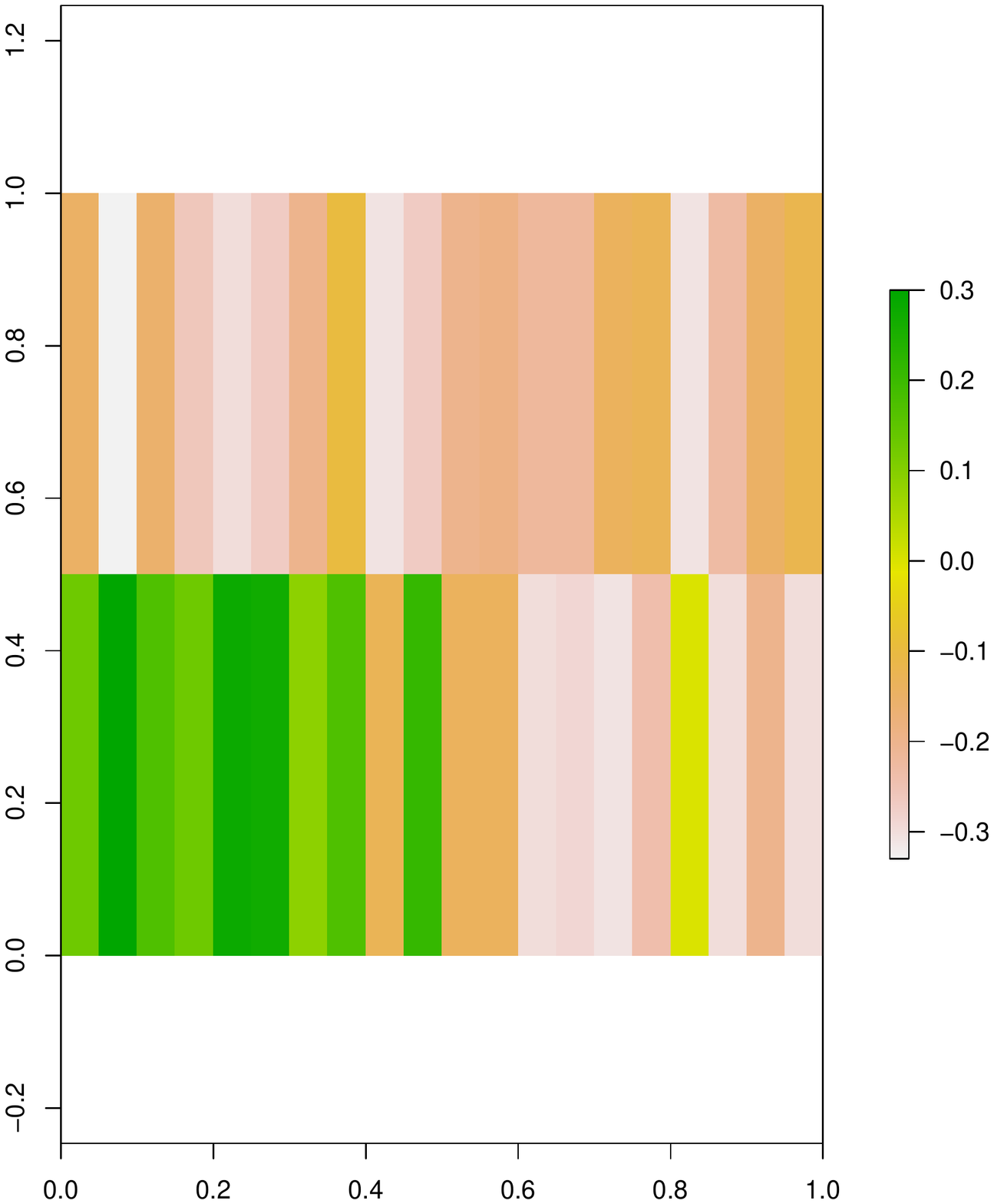}
		\caption{Eigenvectors corresponding to the largest two eigenvalues}
			\label{eigen3}
		%\end{minipage}
	\end{figure}
\begin{figure}[!tbp]
\centering
		%\hfill
		%\begin{minipage}[b]{0.5\textwidth}
			\includegraphics[width=0.4\textwidth]{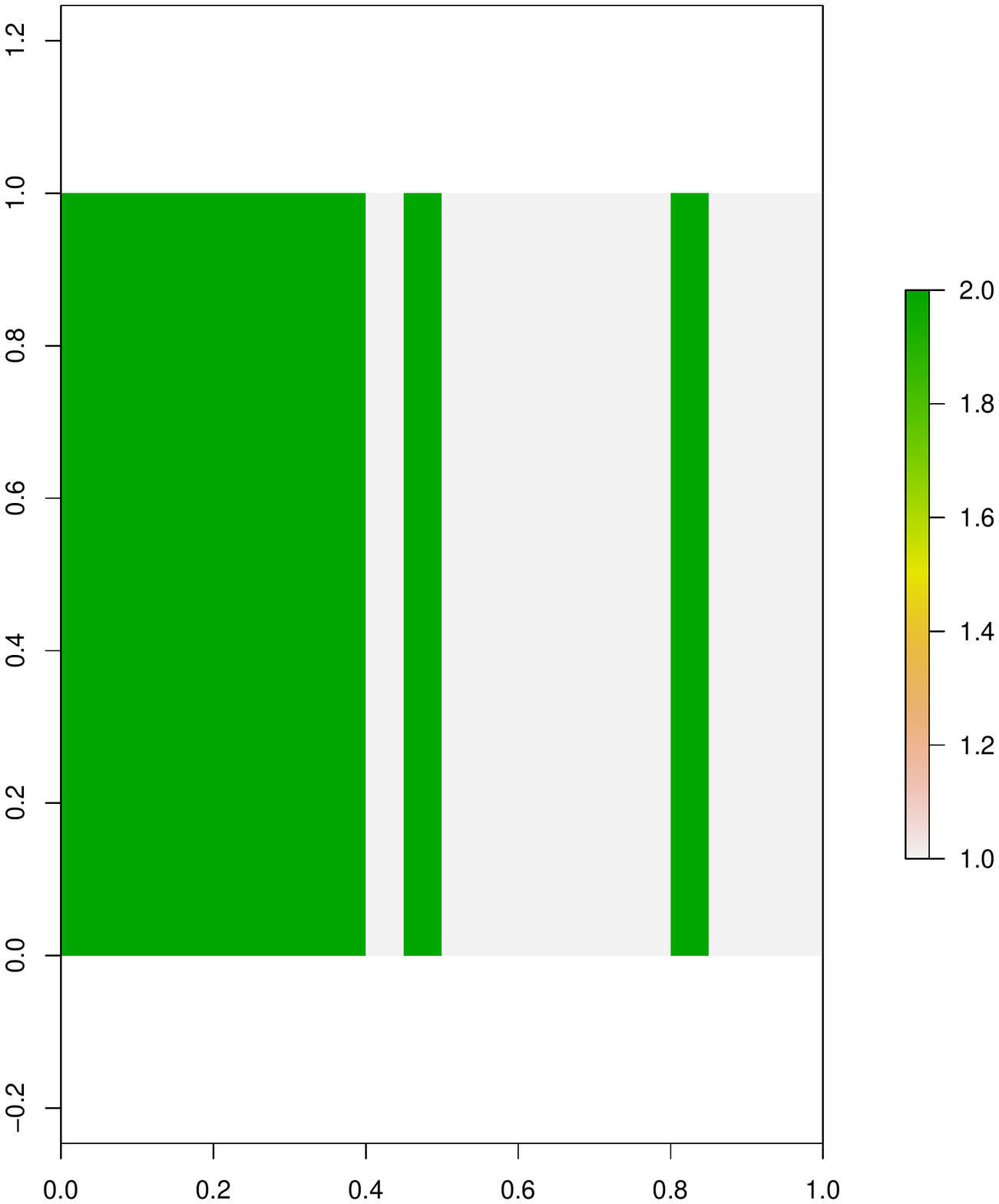}
		\caption{Clustering vector}
			\label{clustervector3}
		%\end{minipage}
	\end{figure}

%
%\begin{table}[ht]
%%\centering
%\caption{Eigenvectors corresponding the largest two eigenvalues}
%	\label{eigen3}
%\begin{lstlisting}[basicstyle=\scriptsize\tt]
%[1] -0.15 -0.33 -0.16 -0.26 -0.30 -0.27 -0.20 -0.10 -0.31 -0.27 
%-0.20 -0.19 -0.22 -0.22 -0.14 -0.13 -0.31 -0.23  -0.15 -0.12
%[2]  0.13  0.30  0.17  0.13  0.28  0.27  0.09  0.17 -0.13  0.21 
%-0.14 -0.14 -0.30 -0.29 -0.31 -0.24  0.00 -0.30  -0.20 -0.30
%		\end{lstlisting}
%		\end{table}
	We can still obtain our desired cluster structure by applying $k$-means algorithm to most rows of  the distance matrix, but some rows give unreliable output. If we use spectral clustering algorithm with respect to the similarity matrix $S$ defined in equation \eqref{similarity}, then we obtain the two eigenvectors of $S$ with the largest two eigenvalues in Figure \ref{eigen3}. Applying $k$-means algorithm to the two eigenvectors in Figure \ref{eigen3}, we obtain the cluster vector in Figure \ref{clustervector3}, which has only 2 misses $(G_9,G_{17})$. 
%\lstinputlisting[float=h,frame=tb,caption=Clustering vector,label=clustervector3 ]{clustervector3.txt}

%\begin{figure}[!tbp]
%		\centering
%		\begin{minipage}[b]{0.3\textwidth}
%			\includegraphics[width=\textwidth]{Clustervector3.pdf}
%		\caption{Clustering vector}
%			\label{clustervector3}
%		\end{minipage}
%	\end{figure}

%\begin{table}[!h]
%	\caption{Clustering vector}
%	\label{clustervector3}
%\begin{lstlisting}
%[1] 2 2 2 2 2 2 2 2 1 2 1 1 1 1 1 1 2 1 1 1
%\end{lstlisting}
%\end{table}

\section{Discussion}
We proposed a novel distance metric between network objects based on the Laplacian flow on graphs by exploiting the long term diffusion behavior of individual networks. With explicit examples, we demonstrated its advantages over various existing distances such as Hamming distance, Frobenius distance between their corresponding graph Laplacians and a maximal diffusion distance. In particular, we find that our network flow distance can detect structure of a network and can be used to accurately classify  or cluster network objects in subsequent clustering tasks.
% especially for those with block structure. 
Therefore it will serve as an important tool for research areas like brain connectomics and shape recognition. Future directions include extending the distance to incorporate  higher-order connectivity of networks by adopting higher-order (Hodge) Laplacians. 

	\section*{Acknowledgments}
	Lizhen Lin was  supported by NSF grants IIS  1663870, DMS 1654579,  DARPA grant N66001-17-1-4041 and an ARO award W911NF-15-1-0440.

%\nocite{*}
\small
%\bibliographystyle{plain}
%%\bibliography{Laplacianbib}

\end{document}